\documentclass[a4paper]{svproc}
\usepackage{url,hyperref}
\usepackage{graphicx}
\usepackage{amsmath}
\usepackage{subcaption} 
\usepackage{booktabs}

\begin{document}
\mainmatter              
\title{Smartwatch-Based Sitting Time Estimation in Real-World Office Settings\thanks{Accepted at the 18th International Conference on Machine Learning and Computing (ICMLC 2026), February 6--9, 2026}}
\titlerunning{Smartwatch-Based Sitting Time Estimation}  
%
\author{Olivia Zhang\inst{1} \and Zhilin Zhang\inst{2}}
%
%
%
\institute{The Hockaday School, Dallas TX 75229, USA\\
\email{ozhang30@hockaday.org},
\and
Lumos Alpha, Dallas TX 75248, USA\\
\email{zhilinzhang@lumosalpha.com}}

\maketitle              

\begin{abstract}
Sedentary behavior poses a major public health risk, being strongly linked to obesity, cardiovascular disease, and other chronic conditions. Accurately estimating sitting time is therefore critical for monitoring and improving individual health.
 This work addresses the problem in real-world office settings, where signals from the inertial measurement units (IMU) on a smartwatch were collected from office workers during their daily routines. We propose a method that estimates sitting time from the IMU signals by introducing the use of rotation vector sequences, derived from Euler angles, as a novel representation of movement dynamics. Experiments on a 34-hour dataset demonstrate that exploiting rotation vector sequences improves algorithm performance, highlighting their potential for robust sitting time estimation in natural environments.

\end{abstract}

\section{Introduction}
\label{sec:intro}

Prolonged sedentary behavior is strongly associated with obesity, cardiovascular disease, metabolic disorders, and mental health conditions \cite{park2020sedentary}. It is the fourth leading risk factor for global mortality, responsible for 6\% of worldwide deaths \cite{park2020sedentary}. It has even been referred to as the “modern smoke.” However, with recent trends in remote working, restrictive urban environments, and long work hours, many office workers find it challenging to reduce their sedentary time.

As a result, estimating sitting duration has become an active research area in wearable computing \cite{hammad2025machine,krauter2024sitting,lui2022apple}. Mekruksavanich et al.\cite{mekruksavanich2018smartwatch} proposed several conventional machine learning algorithms to distinguish sitting from non-sitting activities using smartwatch IMU sensor signals (accelerometer and gyroscope data). They then introduced a hybrid CNN–LSTM model to better capture spatiotemporal dependencies in smartwatch IMU sensor signals \cite{mekruksavanich2020enhanced}. In many such studies, sitting time estimation is often framed as a sitting activity recognition problem, studied within the broader field of human activity recognition \cite{ciortuz2025machine,thakur2021t}. 

Beyond IMU-only methods, alternative approaches have also been explored \cite{salim2024detection,yeon2025watchhar,wang2019deep}. For example, Salim et al.\cite{salim2024detection} proposed a hidden semi-Markov model that relied solely on step counts and heart rate from smartwatch to detect sedentary time. Yeon et al.\cite{yeon2025watchhar} exploited both audio and IMU sensor signals for real-time smartwatch-based activity recognition.

Most of these studies predefine activity types, collect data for each under controlled conditions, and train models to classify movements into one of the predefined categories. However, in real-world settings, the diversity of human activities may make such approaches difficult to generalize. Moreover, many studies have relied on wearable devices placed on the torso, arms, chest, or other body parts \cite{posturemonitorpaper,posturemonitorpaper2,hayashi2021distinguishing,ishii2020exersense}, which are impractical for daily use due to discomfort and inconvenience.


In this work, we address sitting time estimation under real-world conditions by collecting smartwatch IMU data from office workers during their daily routines. Our main contributions are as follows:
\begin{itemize}
  \item We specifically study sitting classification/sitting time estimation using commodity smartwatch IMU signals, rather than relying on other sensing modalities commonly assumed in prior human activity recognition work.
  \item We collect data in real office environments across whole-day routines, where subjects move freely and are not even aware of the ongoing experiment (i.e., without scripted sit/stand instructions and without ``acting for the experiment'' as in prior work), resulting in a noisy dataset containing many micro- and unexpected activities.
  \item Unlike prior studies that rely directly on raw IMU signals \cite{hammad2025machine,krauter2024sitting,lui2022apple,mekruksavanich2018smartwatch,ciortuz2025machine,mekruksavanich2020smartwatch}, we introduce rotation vector sequences, derived from accelerometer data, as an alternative ``signal'' representation that captures orientation-related movement dynamics and provides an additional feature space beyond raw IMU streams. As a result, features extracted from these rotation vector sequences can improve sitting/non-sitting classification performance.
  \item We demonstrate that most existing algorithms can be readily adapted to operate on rotation vector sequences, treating them as an additional sensor modality analogous to accelerometer or gyroscope signals.
\end{itemize}

\section{Proposed Method}
\label{sec:proposed}

We observe that when an office worker is seated at a desk, the natural motion of the arm is constrained, as the desk limits its ability to rest freely at the side of the body. In contrast, when the individual stands, the arm is typically able to hang unobstructed in a relaxed position. In other words, sitting restricts the arm from bending across certain ranges of angles, whereas standing permits greater freedom of movement. Motivated by this observation, we propose the following method for estimating sitting time.

\begin{figure}[t]
  \centering
  \includegraphics[width=0.95\linewidth]{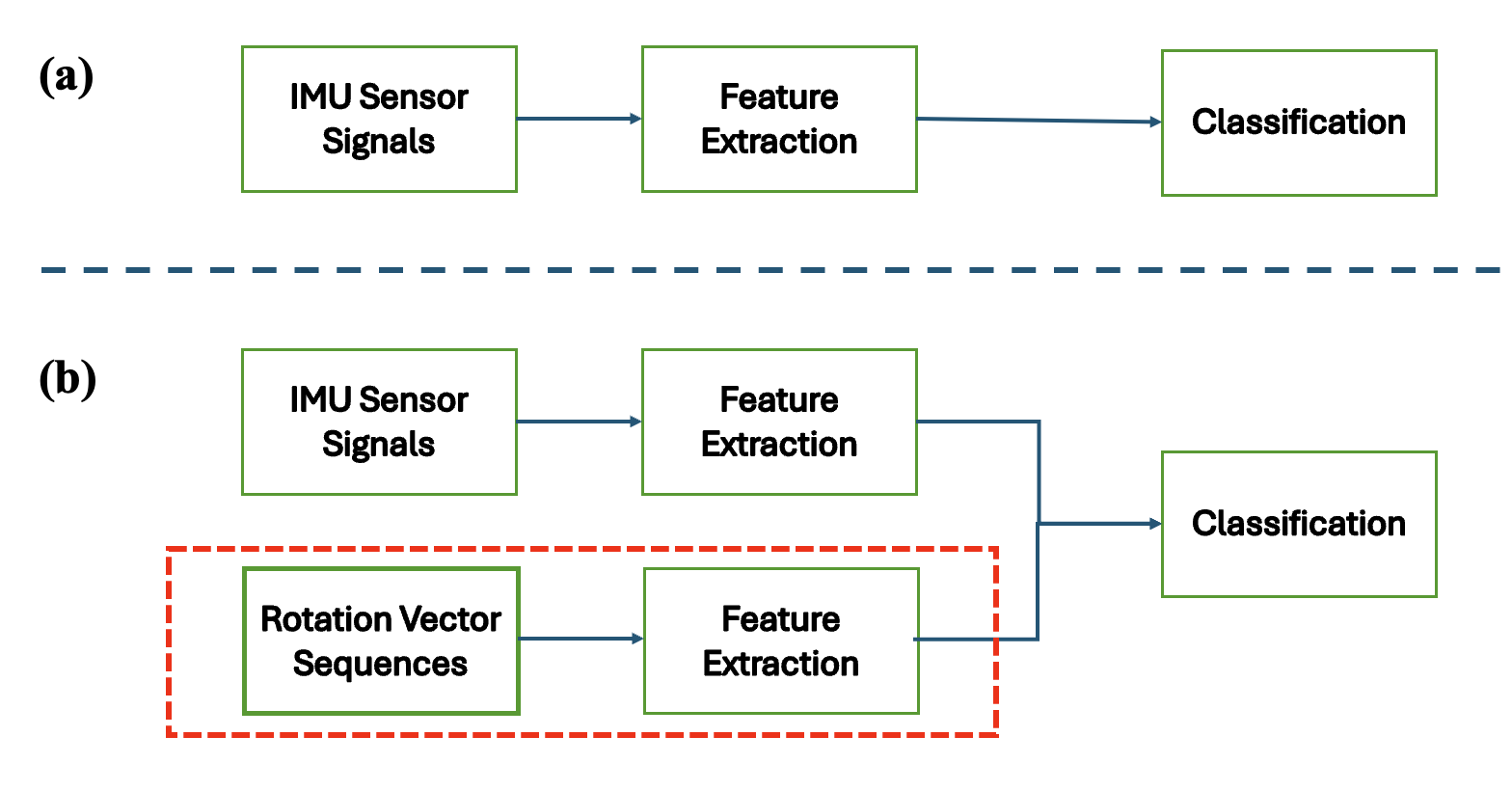} 
  \caption{Comparison between the conventional smartwatch-IMU pipeline and the proposed pipeline. (a) Typical paradigm in prior work for sitting classification/sitting-time estimation, in which features are extracted from IMU signals (e.g., accelerometer and/or gyroscope streams) and fed into a classifier/regressor. (b) Proposed paradigm: in addition to the standard IMU streams, the rotation-vector sequences (indicated by red dashed lines) are treated as an additional signal channel for feature extraction. Features from both branches are fused and then used for classification or regression.}
  \label{fig:framework}
\end{figure}

The proposed method leverages acceleration signals from the smartwatch’s IMU to derive a sequence of rotation vectors. The rotation vector sequence represents movement dynamics of the arm, and thus features derived from these rotation vectors can be utilized to classify sitting versus non-sitting activities. The paradigm of the proposed method is shown in Fig.~\ref{fig:framework}.

\subsection{IMU Coordinate Setting} \label{IMUCoordinate}
In our approach, the smartwatch IMU coordinate system is defined as follows: when the smartwatch lies flat with its screen facing upward, the $x$-axis points to the right, aligned with the wrist’s width; the $y$-axis points forward, aligned with the wrist’s length; and the $z$-axis points outward, perpendicular to the screen. 

Rotation about the $x$-axis is defined as the pitch angle, denoted by $\phi$, whereas rotation about the $y$-axis is defined as the roll angle, denoted by $\theta$. These two angles are the key to our approach. In addition, rotation about the $z$-axis is defined as the yaw angle. We are not interested in this angle because its value can vary arbitrarily regardless of whether an individual is sitting, standing, or walking. Moreover, as we will see soon, without additional information or constraints, the yaw angle cannot be estimated using acceleration signals.

\subsection{Estimate the Pitch and Roll Angles}\label{EstimateAngles}

Note that regardless of what orientation the smartwatch has, gravity is a physical vector fixed in the world frame $\textbf{g}_{\text{world}} = [0, 0, -g]^T$.
When the smartwatch rotates to an orientation, the gravity in its frame can be expressed as $\textbf{g}_{\text{watch}} = [g_x, g_y, g_z]^T$, 
where $g_x$, $g_y$, $g_z$ is the recorded acceleration signal along the axes of the smartwatch.
According to Euler's Theorem \cite{QuaternionsBook}, such a rotation can be achieved via a procedure by first rotating around $z$-axis, then around $y$-axis, and finally around $x$-axis in order.

Assume the smartwatch first rotates around its $z$-axis by a yaw angle $\psi$, the gravity in the smartwatch frame is given by
\begin{equation}
\textbf{g}_{\text{1}} = \textbf{R}_z(\psi) \textbf{g}_{\text{world}}
\label{eq:g1}
\end{equation} 
where $\textbf{R}_z(\psi)$ is a \textbf{passive} rotation matrix around the smartwatch's $z$-axis, given by \footnote{Note that our derivation uses \textbf{passive} rotation matrices (a change of coordinate frame) to express a fixed physical vector, rather than ``spinning'' the vector in 3D space. A passive rotation matrix is the transpose/inverse of the corresponding \textbf{active} rotation matrix, which is commonly seen in the literature \cite{QuaternionsBook}.}
\[
\mathbf{R}_z(\psi) =
\begin{bmatrix}
\cos\psi & \sin\psi & 0 \\
-\sin\psi & \cos\psi & 0 \\
0 & 0 & 1
\end{bmatrix}
\]

Next, the smartwatch rotates around its $y$-axis by a roll angle $\theta$. The gravity in the smartwatch's frame is then
\begin{equation}
\textbf{g}_{\text{2}} = \textbf{R}_y(\theta) \textbf{g}_{\text{1}}
\label{eq:g2}
\end{equation} 
where $\textbf{R}_y(\theta)$ is a passive rotation matrix around the $y$-axis, given by
\[
\textbf{R}_y(\theta) =
\begin{bmatrix}
\cos\theta & 0 & -\sin\theta \\
0 & 1 & 0 \\
\sin\theta & 0 & \cos\theta
\end{bmatrix}
\]

Finally, the smartwatch rotates around its $x$-axis by a pitch angle $\phi$. At this step, the gravity in the smartwatch's frame is
\begin{equation}
\textbf{g}_{\text{watch}} = \textbf{R}_x(\phi) \textbf{g}_{\text{2}}
\label{eq:g3}
\end{equation} 
where $\textbf{R}_x(\phi)$ is the passive rotation matrix around the $x$-axis, given by
\[
\textbf{R}_x(\phi) =
\begin{bmatrix}
1 & 0 & 0 \\
0 & \cos\phi & \sin\phi \\
0 & -\sin\phi & \cos\phi
\end{bmatrix}
\]

Combining the three steps (\ref{eq:g1}) - (\ref{eq:g3}), we have
\begin{align}
\mathbf{g}_{\text{watch}}
&= \mathbf{R}_x(\phi)\,\mathbf{R}_y(\theta)\,\mathbf{R}_z(\psi)\,\mathbf{g}_{\text{world}} \nonumber\\[4pt]
&=
\begin{bmatrix}
1 & 0 & 0 \\
0 & \cos\phi & \sin\phi \\
0 & -\sin\phi & \cos\phi
\end{bmatrix}
\begin{bmatrix}
\cos\theta & 0 & -\sin\theta \\
0 & 1 & 0 \\
\sin\theta & 0 & \cos\theta
\end{bmatrix} \nonumber\\[-2pt]
&\quad\;
\begin{bmatrix}
\cos\psi & \sin\psi & 0 \\
-\sin\psi & \cos\psi & 0 \\
0 & 0 & 1
\end{bmatrix}
\begin{bmatrix}
0\\[2pt]
0\\[2pt]
-g
\end{bmatrix} \nonumber\\[6pt]
&=
\begin{bmatrix}
g\sin\theta \\
-\,g\sin\phi\,\cos\theta \\
-\,g\cos\phi\,\cos\theta
\end{bmatrix}. \label{eq_gwatch}
\end{align}
Note that the yaw angle $\psi$ does not appear in the final result. This is why the yaw angle cannot be estimated without using additional information. Fortunately, as stated before, we are not interested in estimating the yaw angle.

From \eqref{eq_gwatch} we have
\begin{align}
g_x &= g \sin\theta \\
g_y &= -g \sin\phi \cos\theta \\
g_z &= -g \cos\phi \cos\theta
\end{align}
From there, it is simple to solve for $\phi$ and $\theta$:
\begin{align}
\phi   &= \operatorname{atan2}(-g_y,-g_z) \\
\theta &= \operatorname{atan2}\!\bigl(g_x,\sqrt{g_y^2+g_z^2}\bigr)
\end{align}
The estimated angles $\phi$ and $\theta$ exhibit distinct patterns between sitting and non\text{-}sitting postures and can therefore serve as features for classification. However, the solutions are susceptible to a phenomenon known as \textbf{gimbal lock}: when $|\theta|$ approaches $90^{\circ}$ (i.e., $\cos\theta \to 0$), $\phi$ becomes ill-defined and numerically unstable. As a result, using $\phi$ and $\theta$ directly as features is unreliable near the singularity and should be avoided.

\subsection{Rotation Matrix and Axis-Angle Representation}

Instead of using $\phi$ and $\theta$ directly as features, we consider their corresponding rotation matrices. Defining $\textbf{R}_{xy} = \textbf{R}_x(\phi) \textbf{R}_y(\theta)$, we have
\begin{align}
\textbf{R}_{xy}
&=
\begin{bmatrix}
1 & 0 & 0 \\
0 & \cos\phi & \sin\phi \\
0 & -\sin\phi & \cos\phi
\end{bmatrix}
\begin{bmatrix}
\cos\theta & 0 & -\sin\theta \\
0 & 1 & 0 \\
\sin\theta & 0 & \cos\theta
\end{bmatrix} \nonumber \\
&= \begin{bmatrix}
\cos\theta & 0 & -\sin\theta \\
\sin\phi\,\sin\theta & \cos\phi & \sin\phi\,\cos\theta \\
\cos\phi\,\sin\theta & -\sin\phi & \cos\phi\,\cos\theta
\end{bmatrix}. \label{eq:comrotation}
\end{align}
It should be noted that although $\phi$ and $\theta$ may have gimbal lock, the combined rotation matrix $\textbf{R}_{xy}$ evolves smoothly. Therefore, we design features from the rotation matrix \eqref{eq:comrotation}.

The $3 \times 3$ combined rotation matrix has orthogonality constraints. To better capture its evolvement pattern, one can derive the rotation vector associated with it, which is a compact, unconstrained 3D representation that respects the geometry of rotations and is easier for learning algorithms.

Following the approach in \cite{rotmatwiki}, $\textbf{R}_{xy}$ can be expressed as a rotation by angle $\alpha \in [0, \pi]$ about a unit axis $\mathbf{u} \in \mathbf{R}^3$, where
\[
\alpha \;=\; \arccos\!\left(\frac{\operatorname{tr}(\textbf{R}_{xy})-1}{2}\right),
\]
and, for $\alpha \neq 0$,
\[
\mathbf{u}
\;=\;
\frac{1}{2\sin\alpha}
\begin{bmatrix}
R_{32} - R_{23}\\[2pt]
R_{13} - R_{31}\\[2pt]
R_{21} - R_{12}
\end{bmatrix}
\]
where $R_{ij}$ is the $(i,j)$-th element of $\textbf{R}_{xy}$. 

The rotation vector is then
\[
\mathbf{r} \;=\; \alpha\,\mathbf{u} 
\;=\;
\frac{\alpha}{2\sin\alpha}
\begin{bmatrix}
R_{32} - R_{23}\\[2pt]
R_{13} - R_{31}\\[2pt]
R_{21} - R_{12}
\end{bmatrix}.
\]
When $\alpha \approx 0$, using the first-order approximation, we have
\[
\mathbf{r}
\;\approx\;
\begin{bmatrix}
\frac{R_{32} - R_{23}}{2}\\[2pt]
\frac{R_{13} - R_{31}}{2}\\[2pt]
\frac{R_{21} - R_{12}}{2}
\end{bmatrix}.
\]

The rotation vector $\mathbf{r}$ represents the final output. At each timestamp, a rotation vector can be computed from the three-axis acceleration signals. Consequently, a temporal sequence of acceleration measurements can be transformed into a corresponding sequence of rotation vectors, from which discriminative features may be extracted for classification tasks. Alternatively, deep learning models can be applied directly to the sequence of rotation vectors, thereby circumventing reliance on raw acceleration signals.

To exploit the rotation vectors, it is essential that the acceleration signals measured by the smartwatch IMU, $\mathbf{g}_{\text{watch}}$, are dominated by the gravitational component, while contributions from dynamic accelerations due to external forces are minimized. Depending on the application, this requirement may be relaxed, or it can be addressed through signal processing techniques such as low-pass filtering, smoothing, or sensor fusion with gyroscope data. In the next section, we demonstrate how the sequence of rotation vectors can be leveraged for the sitting time estimation application.

It is important to note that the majority of existing studies \cite{hammad2025machine,mekruksavanich2018smartwatch,salim2024detection} have focused on directly using acceleration signals for human activity classification. In contrast, our approach introduces a novel paradigm by transforming acceleration signals into rotation vectors, from which discriminative features can be derived or end-to-end models can be trained (See Fig.~\ref{fig:framework}). This representation captures orientation information more effectively and opens a new, potentially robust direction for advancing human activity recognition research.


\section{Classification}

A variety of approaches can be employed to exploit the sequence of rotation vectors for distinguishing sitting from non-sitting states. In this work, we propose a CatBoost-based algorithm \cite{prokhorenkova2018catboost}. More importantly, we emphasize that most existing methods can be enhanced by treating the rotation vector sequence as an additional sensor modality and applying algorithms to it in the same way they are applied to acceleration or gyroscope signals.

Following prior work \cite{hammad2025machine,mekruksavanich2018smartwatch}, we segment the tri-axial accelerometer time series into windows of length \(N\). For each window, we compute the median of the \(x\)-, \(y\)-, and \(z\)-axis signals. When the watch-wearing arm is largely stationary (e.g., reading, writing, typing), these medians closely approximate the gravitational components along each sensor axis, enabling reliable estimation of the device’s orientation and the corresponding rotation vector. During vigorous activites (e.g., walking or running), however, linear accelerations dominate and the readings from the accelerometer, $\textbf{g}_{watch}$, will not entirely be based off of gravity. But we find the rotation vector sequence, calculated at each timestamp, exhibits evolution patterns indicating of non-sitting behavior. Therefore, in addition to the axis-wise median-derived rotation vector, we calculate the following features from the rotation vector sequence: $\mathbf{r}^{n,m} = [r_x^{n,m},r_y^{n,m},r_z^{n,m} ]^T$, where $n \in \{0,\dots,N-1\}$ indexes the timestamps within the window, and $m$ denotes the window index.

One type of important feature is the Fuzzy entropy \cite{chen2007characterization} of $r_x^{n,m}$, $r_y^{n,m}$, and $r_z^{n,m}$. This complexity measure quantifies the irregularity and unpredictability of a time series. Unlike Shannon entropy and other conventional entropy measures, fuzzy entropy is specifically designed for time-series analysis, providing more robust and consistent characterization of temporal dynamics.

Another type of feature is the statistical descriptors of $r_x^{n,m}$, $r_y^{n,m}$, and $r_z^{n,m}$, 
    including the mean, standard deviation, maximum, minimum, interquartile range (IQR), and median absolute deviation (MAD). These features capture the central tendency, variability, and distributional spread of a time series.

The energy measure is also used, defined for $r_x^{n,m}$ as $E_x = \frac{1}{N} \sum_{n=0}^{N-1} \bigl(r_x^{n,m}\bigr)^2$ with analogous definitions for $r_y^{n,m}$ and $r_z^{n,m}$.

In addition to extracting the above features from the rotation vector sequence $\mathbf{r}^{n,m}$ to characterize its temporal evolution patterns, we also compute the same set of features directly from the raw acceleration signals (and gyroscope signals if available). This dual representation enables the model to exploit both orientation-related information from the rotation vectors and raw signal characteristics, thereby improving the robustness and accuracy of classification.

Furthermore, the sequence of rotation vectors can be viewed as another sensor modality, providing an alternative representation of movement dynamics. Most existing algorithms can be directly applied to the sequence of rotation vectors with improved performance, just as they are applied to tri-axial acceleration and gyroscope signals. The experimental results presented in Section~\ref{sec:experiments} demonstrate this effect.


\section{Experiments}
\label{sec:experiments}

Our study aims to estimate sitting time for office workers under real-world, uncontrolled conditions. To this end, we collected tri-axial accelerometer signals (and gyroscope signals) from a smartwatch IMU worn on the non-dominant wrist of five office workers during their regular workdays. Each participant wore the smartwatch for at least five consecutive hours while performing routine office activities, such as sitting while working on  a computer, sitting to have a lunch, making coffee, talking with colleagues in the hallway, attending meetings, and engaging in group discussions at a whiteboard. One of the authors accompanied each participant as a visitor to their workplace and manually recorded sitting and non-sitting timestamps. (Video recording was not permitted due to company privacy policies.) In total, the dataset comprises over 34 hours of recordings sampled at 100 Hz, with annotations of sitting and non-sitting intervals.

Following the general practices in the field \cite{hammad2025machine,mekruksavanich2018smartwatch}, the raw data were segmented into windows of 30 seconds. Although such a window length may be considered long in some human activity recognition tasks, it is appropriate for sitting time estimation. For each window, if sitting time accounted for more than half of the duration, the signals within that window were labeled as sitting; otherwise, they were labeled as non-sitting. In this way, the sitting time estimation problem is reformulated as a binary classification task.

To evaluate classification performance, we employed standard metrics, including accuracy, precision, recall, and F1-score with respect to the sitting class. We compared our proposed algorithm with several representative algorithms: the ST-DT algorithm by Mekruksavanich et al.\cite{mekruksavanich2018smartwatch}, which achieved the best performance in their sitting detection study; the hybrid CNN–LSTM model \cite{mekruksavanich2020enhanced}; the variational autoencoder (VAE) approach \cite{fan2020new}; the MaskCAE algorithm \cite{cheng2024maskcae}; and the Conformer-based method \cite{seenath2023conformer}. In addition, the compared algorithms were slightly modified to extend their applicability to the sequence of rotation vectors, thereby demonstrating the generalizability of rotation vector–based representations across different modeling approaches.

All algorithms were trained on 80\% of randomly selected samples and evaluated on the remaining 20\%. A 5-fold cross-validation was applied. The results are summarized in Table \ref{tab:exp_comparison}, from which we can see the proposed method achieved the highest performance among all algorithms. More importantly, all compared algorithms demonstrated improved performance when both raw IMU sensor signals and rotation vector sequences were utilized, highlighting the benefits of incorporating rotation vector representations for sitting time estimation.

\begin{table}[t]
\centering
\caption{Performance comparison of the proposed method with ST-DT, CNN-LSTM, VAE, MaskCAE, and Conformer. The variants adapted to utilize rotation vector sequences are denoted as ST-DT Improved, CNN-LSTM Improved, VAE Improved, MaskCAE Improved, and Conformer Improved. Recall, precision, and F1-score are reported with respect to the sitting class.}

\label{tab:exp_comparison}
\renewcommand{\arraystretch}{1} 
\setlength{\tabcolsep}{2.8pt}       
\small                       
\begin{tabular}{lcccc}
\toprule
\textbf{Method} & \textbf{Recall} & \textbf{Precision} & \textbf{F1-score} & \textbf{Accuracy} \\
\midrule
\textbf{Proposed}  & \textbf{0.964} & \textbf{0.978} & \textbf{0.971} & \textbf{0.958} \\
ST-DT               & 0.905 & 0.947 & 0.925 & 0.894 \\
\textbf{ST-DT Improved}      & \textbf{0.928} & \textbf{0.938} & \textbf{0.933} & \textbf{0.904} \\
CNN-LSTM                      & 0.896 & 0.967 & 0.930 & 0.902 \\
\textbf{CNN-LSTM Improved}    & \textbf{0.921} & \textbf{0.981} & \textbf{0.950} & \textbf{0.930} \\
VAE                           & 0.916 & 0.974 & 0.944 & 0.922 \\
\textbf{VAE Improved}         & \textbf{0.916} & \textbf{0.989} & \textbf{0.951} & \textbf{0.932} \\
MaskCAE                      & 0.926 & 0.981 & 0.953 & 0.933 \\
\textbf{MaskCAE Improved}    & \textbf{0.928} & \textbf{0.984} & \textbf{0.955} & \textbf{0.937} \\
Conformer                    & 0.972 & 0.955 & 0.963 & 0.947 \\
\textbf{Conformer Improved}  & \textbf{0.957} & \textbf{0.977} & \textbf{0.967} & \textbf{0.952} \\
\bottomrule
\end{tabular}
\end{table}


\section{Conclusion}

Smartwatch-based sitting time estimation is an emerging trend in wearable computing and is increasingly becoming a built-in functionality of modern smartwatches. While prior studies have investigated this problem, most have relied on data collected in controlled laboratory settings where subjects were instructed to perform repeated sitting or non-sitting actions.

In contrast, this work addresses the problem under real-world conditions by collecting data from office workers during their daily routines in the workplace. We proposed an algorithm for sitting time estimation based on sequences of rotation vectors derived from tri-axial acceleration signals. Our study demonstrates that these rotation vector sequences serve as an alternative representation of movement dynamics, capturing informative patterns that can be effectively exploited to improve classification performance. More importantly, this representation can be leveraged by most existing algorithms with only minor modifications, allowing them to operate on rotation vector sequences as naturally as on tri-axial sensor signals. Experimental results confirm that incorporating these representations enhances sitting time estimation.

For future work, we plan to validate the proposed rotation-vector representation on larger and more diverse real-world datasets (e.g., varied job types, arm/wrist habits, and environments) to better assess generalization beyond the current 34-hour office collection. We also aim to improve robustness under vigorous motion by incorporating additional sensor signals and investigating the temporal dynamics exhibited in rotation-vector sequences.


%
%

\end{document}